\documentclass[twocolumn]{article}
\usepackage{amsmath}
\usepackage{amssymb}
\usepackage{geometry}
\usepackage{booktabs}
\usepackage{appendix}
\usepackage{tikz}
\usepackage{multicol}
\usepackage[utf8]{inputenc}

\title{Using language models in the implicit automated assessment of mathematical short answer items}
\author{Christopher Ormerod\\ \\
Cambium Assessment \\
1000 Thomas Jefferson St NW, Washington\\
District of Columbia, 20007, United States\\ \\ 
christopher.ormerod@cambiumassessment.com
}

\begin{document}

\maketitle

\begin{abstract}
We propose a new way to assess certain short constructed responses to mathematics items. Our approach uses a pipeline that identifies the key values specified by the student in their response. This allows us to determine the correctness of the response, as well as identify any misconceptions. The information from the value identification pipeline can then be used to provide feedback to the teacher and student. The value identification pipeline consists of two fine-tuned language models. The first model determines if a value is implicit in the student response. The second model identifies where in the response the key value is specified. We consider both a generic model that can be used for any prompt and value, as well as models that are specific to each prompt and value. The value identification pipeline is a more accurate and informative way to assess short constructed responses than traditional rubric-based scoring. It can be used to provide more targeted feedback to students, which can help them improve their understanding of mathematics.
\end{abstract}

\section{Introduction}

Automated short answer grading (ASAG) is the use of statistical models to assess student short constructed responses in a way that is designed to replicate human scoring \cite{leacock_c-rater_2003}. ASAG is often seen as an ideal application of artificial intelligence because the rubrics for short constructed responses are relatively simple and the process of hand-scoring them can be time-consuming. However, ASAG has not received as much attention as automated essay scoring (AES) \cite{ramesh_automated_2022, haller_survey_2022}. One of the most challenging areas of ASAG is the assessment of items in mathematics. This is because mathematics questions often require students to demonstrate their understanding of concepts and procedures, which can be difficult for a machine to evaluate \cite{baral_improving_nodate, zhang_automatic_2022}. 

Transformer-based language models are a type of deep learning model \cite{vaswani_attention_2017} that have achieved state-of-the-art results on a wide range of natural language processing tasks, such as question answering, natural language inference, and text summarization \cite{wang_glue_2019, wang_superglue_2020}. They are distinguished by their use of the attention mechanism, which allows them to learn long-range dependencies between words in a sentence. Pretraining is a technique used in natural language processing to train a language model on a large amount of unlabeled text data. This allows the model to learn general language understanding and generation skills, which can then be applied to a specific task \cite{devlin_bert_2019}. Pretrained language models (PLMs) have been shown to be effective for a variety of tasks, including automated essay scoring (AES) and automated short answer grading (ASAG). In AES, PLMs can be used to score essays by identifying key features of good writing, such as grammar, vocabulary, and coherence. In ASAG, PLMs can be used to score short constructed responses by identifying key concepts and procedures. Recent studies have shown that PLMs can achieve state-of-the-art results on AES \cite{uto_neural_2020} and ASAG tasks \cite{ormerod_automated_2022}. 

Despite all the advances with PLMs in ASAG, mathematical reasoning has proven to be difficult even for Large Generative Language Models \cite{floridi_gpt-3_2020}. This challenge has prompted researchers to develop models that are specific to mathematics \cite{peng_mathbert_2021}. The fundamental challenge presented by mathematical reasoning tasks requires developing a multitude of different ways to approach the automated assessment of items in mathematics. 

This short article concerns a method of ASAG where the assessment of the item is approached implicitly instead of explicitly. Instead of assigning a score based on a holistic rubric, we use language models to extract key numerical values from the response. Not only does this information allow us to assign a score, these values are able to highlight some of the misconceptions the student may have. This addresses the growing need in education for language models to provide feedback to teachers and students \cite{mcnichols_exploring_2023}.

The dataset we used for this task was derived as a supplementary dataset provided in connection with a national assessment program. It was administered alongside a short answer dataset used in a previous paper \cite{ormerod_short-answer_2022}. This system relies upon extensive preprocessing to convert and normalize the set of numerical values that appear in a constructed response and a pipeline consisting of a classification model and a question answering model. We use language models for both of these modeling tasks. 

This article is organized as follows: in \S \ref{sec:method}, we give a brief description of the dataset used in this study, how we preprocessed responses, and details regarding how the dataset was modeled. We then consider a comparison of the engines results against human benchmarks in \S \ref{sec:results}. In \S \ref{sec:conclusion}, we discuss the results and future directions for this research.

\section{Method}\label{sec:method}

\subsection{Approach}

The type of question we wish to address is one in which the student is prompted to present a collection of values, the combination of which constitutes a correct answer. The nature of these questions is perhaps best explained by presenting an example of such a question:

\begin{quote}
{\em 
Question: There are 3 types of bags of candies for purchase for Halloween. Bags of chocolates are \$7 each, lollipops are \$3 each, and gum sticks are \$5 each. Give one way we can spend a total of \$64 using a combination of chocolates, lollipops, and gum sticks.}
\end{quote}

While the question reflects the characteristics of the items being considered, it is not one of the prompts associated with the dataset used in this study. The actual questions used in this study have copyright agreements that prevent their public dissemination. 

Any response needs to specify a collection of quantities, and if the quantity is absent from the response, it is assumed to be 0. To present a possible response, we used the Llama 2 model with 70 billion parameters \cite{touvron_llama_2023}, which produced the following:

\begin{quote}
{\em One possible way to spend a total of \$64 using a combination of chocolates, lollipops, and gum sticks would be to purchase 9 bags of chocolates ($\$7\times 9 = \$ 63$), 1 bag of lollipops ($\$ 3$), and 2 bags of gum sticks ($\$5 \times 2 = \$ 10$). This combination would total \$64.}
\end{quote}

To be clear, ChatGPT has been shown to be unreliable in providing valid solutions to this problem \cite{floridi_gpt-3_2020}. We believe that this is because numbers are not always well distinguished by their contextual semantic embedding. 

One of the difficulties is that there are multiple solutions. A simple calculation reveals there are 24 distinct solutions to this problem. Each of these solutions can be expressed in a variety of ways, making it difficult for a machine to determine whether a student's response is correct. One approach to this challenge is to train a language model on a large corpus of correct solutions. This would allow the model to learn to identify the key features of a correct solution, even if it is expressed in a different way. However, this approach is limited by the availability of correct solutions. In many cases, there may not be enough correct solutions to train a reliable model.

Another challenge is that ASAG systems need to be robust to misspelled words and incorrect grammar. This is because spelling and grammar are often not part of the rubric for ASAG. For example, a student may correctly solve a math problem but make a spelling error in their response. An ASAG system should be able to overlook these errors and still give the student a correct score.

It is also true that we use some form of question answering to ascertain the number of bags of chocolates, which we call $x$, the number of bags of lollipops, $y$, and the number of gum sticks, $z$, then it is easy to check that we have a valid solution. We need only check that $x,y,z \in \mathbb{N}$ and the condition $7x+3y+5z = 64$ is satisfied. Our ability to ascertain valid solutions is contingent upon our ability to accurately determine $x$, $y$, and $z$.

The numerical information in student responses is valuable for more than just determining whether the responses are correct. It can also be used to highlight misconceptions and provide feedback. For example, in the above example, we can determine whether the student has spent over or under the required amount. Similarly, in a question regarding fractions, the numerator and denominator can tell us whether the student understands common denominators or whether the student has trouble simplifying fractions. This information can be used to provide feedback on the student's understanding of fraction operations.

The pipeline described in this paper is designed to simplify the extraction of these numerical values from student responses. This can make it easier for teachers and educational researchers to identify and address misconceptions in student learning.

\subsection{Data}

The dataset used in this study is associated with 7 prompts, each with 1 to 12 values to extract. For each prompt, there are either 4,000 or 8,000 responses that were double scored. The human scorers were instructed to read each response and determine whether the specific value was stated in the text. If the value was stated, they would enter the value in the data associated with that response, and if it was not, the rater left the value blank. Unlike Question Answering (QA) datasets like SQuAD \cite{rajpurkar_squad_2016}, the location of the value in the text is not recorded. If two raters disagreed, the final score was resolved by an expert. These scoring processes were the same used in a previous study \cite{ormerod_automated_2022}.

The pipeline we use consists of two models: a text classification model and a token classification model. The text classification model distinguishes between three types of answers: 0, 1, or a value, $v$, which is any other value mentioned in the text. This distinction is necessary because there are many ways to specify a value of zero or one in words, without explicitly stating a number. However, there are very few examples of values other than zero or one being specified without explicitly stating the number. The distribution of resolved classes for the complete dataset can be found in Table \ref{tab:classes}.

\begin{table}
\begin{tabular}{c c| c | c c c  } \toprule
Prompt & V & N  & $0$ & $1$ & $v$ \\ \midrule
1 & 9 & 8k & 87.2 & 2.6 & 10.2 \\
2 & 3 & 4k & 59.4 & 4.6 & 36.0 \\
3 & 1 & 4k & 52.0 & 0.0 & 48.0 \\
4 & 1 & 4k & 24.9 & 0.2 & 74.9 \\
5 & 1 & 4k & 18.6 & 0.5 & 80.9 \\
6 & 8 & 8k & 65.3 & 17.8 & 16.8 \\
7 & 12 & 4k & 65.1 & 2.2 & 32.8 \\ \bottomrule
\end{tabular}
\caption{The distribution of values of 0, 1, or $v$ where $v$ is any other value. The value of $N$ represents the number of responses and $V$ is the number of values to extract from each prompt. \label{tab:classes}}
\end{table}

Given we have double scored responses, the level of agreement between two raters can be evaluated using Cohen's kappa score \cite{cohen_coefficient_1960}, given by
\[
\kappa = \dfrac{p_o - p_e}{1-p_e}
\]
where $p_o$ is the observed probability and $p_e$ is the expected probability. This measure is also fairly standard within the industry more broadly \cite{williamson_framework_2012}.

For each prompt, and for the entire dataset, we define $\kappa_0$ to be the kappa value indicating that both raters agree that either no value has been specified or the student has specified a value of 0. Similarly, we define $\kappa_1$ to be the kappa value for the agreement that the student has specified a value of 1, and $\kappa_v$ is the agreement that any other value has been specified in the text. A summary of the inter-rater reliability for the data used in this classification problem is found in Table \ref{tab:irr_classification}. To simplify the presentation, we treat each value in each prompt independently, hence, given a prompt with $N$ responses and $V$ values, the kappa values presented in Table \ref{tab:irr_classification} are calculated over $N\times V$ different cases. 

\begin{table}[!ht]
    \centering
    \begin{tabular}{c c| c | c c c  } \toprule
Prompt & V & N  & $\kappa_0$ & $\kappa_1$ & $\kappa_v$ \\ \midrule 
1 & 9 & 8k & 0.866 & 0.877 & 0.853 \\
2 & 3 & 4k & 0.910 & 0.899 & 0.920 \\
3 & 1 & 4k & 0.782 & 1.000 & 0.782 \\
4 & 1 & 4k & 0.795 & 1.000 & 0.795 \\
5 & 1 & 4k & 0.714 & 0.897 & 0.724 \\
6 & 8 & 8k & 0.903 & 0.904 & 0.893 \\
7 & 12 & 4k & 0.932 & 0.938 & 0.935 \\
Total & 35 & 40k & 0.906 & 0.910 & 0.906 \\ \bottomrule
    \end{tabular}
    \caption{The inter-rater reliability indicating when the raters agreed that the value being specified by the student was 0, 1, or some other value present in the text.}
    \label{tab:irr_classification}
\end{table}

The second model is used in the situation associated with $\kappa_v$, where a value is specified in the response. From the entire set of responses, the data for the second model consists of all responses in which the value is found to be specified somewhere in the text. If this is the case, at least one of the raters would have found a value, in which case, the probability of exact match between raters, $p$, is a valid metric to measure agreement. These $p$ values are presented in Table \ref{tab:irr_acc}.

\begin{table}[!ht]
    \centering
\begin{tabular}{r rrrrr}
\toprule
Prompt & V & min & max & avg & p \\
\midrule

1 & 9  & 242 & 1903 & 1104 & 0.844 \\
2 & 3  & 857 & 3389 & 1701 & 0.913 \\
3 & 1  & - & - & 2043 & 0.852 \\
4 & 1  & - & - & 3144 & 0.920 \\
5 & 1  & - & - & 3461 & 0.915 \\
6 & 8  & 1821 & 5311 & 3005 & 0.897 \\
7 & 12 &  328 & 2479 & 1429 & 0.930 \\
\bottomrule
\end{tabular}
    \caption{The various prompts alongside the number of values, $V$, to be extracted. We also present the minimum, maximum, and average number of training examples to train the model used to extract values.}
    \label{tab:irr_acc}
\end{table}

This dataset was partitioned into a random set of training responses consisting of exactly 70\% of all data, a development set consisting of 15\% of all data, and a test set consisting of the remaining 15\% of all responses. The evaluation of a pipeline, involving both the classification and identification models, is given by the tuple $(\kappa_0, \kappa_1, \kappa_v, p)$ measured on the test set. The development set is used for hyperparameter tuning and model choice. 

One of difficulties presented by this dataset is that there are too few samples to train specific models for each value we wish to extract. One prompt contains a value that is only specified by the student approximately 3\% of the time. This means that the expected number of training examples would be 169. This is far too few to fine tune a model. 

\subsection{Preprocessing}

When extracting values from written text, we need to address the issue that there are a number of equivalent ways to express a number. We assume students write numbers in either decimal, (improper) fractional, or written form (ignoring esoteric representations such as Roman numerals, binary, or hexadecimal). We can use regular expressions to augment text so that we provide a standardized decimal or simplified fractional form alongside any number the student provides for the models to choose if appropriate. This process of normalizing equivalent forms of numerical data helps the engine identify equivalent values. 

The most difficult part of the preprocessing was to identify numbers that have been specified in words within the text. Fortunately, the set of words used to specify most numbers is small. Additionally, the many words specifying numbers can be used interchangeably. This allows us to design a finite state automata (FSA) to determine whether a sequence of words specifies a valid number. The FSA can be used iteratively to identify written representations of words and convert those representations to floating point representation of their corresponding numerical values.

\subsection{Modeling}

The first modeling task considered the classification of the student response as either being 0, 1, or $v$ where $v$ is any particular value specified by the student. This can be treated as a standard text-classification task. As shown in Table \ref{tab:classes}, the data for this classification task is highly imbalanced, with the values of 1 only being assigned on average approximately 7.1\% of the time. For prompt 3, there are so few that the overall percentage rounds down to 0. If we consider models tuned to specific values in a prompt, there are far too few instances of this class for some prompts and values for any meaningful modeling. 

For this reason, we consider generic models, which can assess a pair consisting of a question and a student response. The idea is that generic models are able to use patterns learned from other prompts and values to make inferences. For such a generic model, we format our input as follows:
\begin{verbatim}
<cls><prompt><sep><response><sep>
\end{verbatim}
where \verb_<sep>_ and \verb_<cls>_ are special tokens used by the language model to separate segments of text. 

Since this is a standard classification task, we choose a small version of the Efficiently Learning an Encoder that Classifies Token Replacements Accurately (ELECTRA) model \cite{clark_electra_2020}. The pretraining objective was to determine which tokens were replaced by a corresponding generator model. This training objective is different to the one used to train masked language models such as BERT \cite{devlin_bert_2019}. The small ELECTRA discriminator model is a very small model, with just 13 million parameters, while maintaining excellent downstream performance in GLUE benchmarks \cite{wang_glue_2019}. It has been used in a variety of educational applications \cite{ormerod_automated_2021}.

The second model is considered a token classification model. It is also the case that some prompt-value pairs do not have sufficiently many training examples to fine-tune a single model, hence, generic models are also used to identify values specified by the students. 

In considering generic models, we found that a single model did not provide adequate and consistent performance across prompts and values. That is to say that a single model often provided good performance on one task, but failed in others. The inconsistent results of a single model could possibly be attributed to the very small training set for some particular values. In the end, we considered an ensemble of an ELECTRA model, described above, and a version of BERT, called Math-BERT, which was trained on a mathematical dataset \cite{peng_mathbert_2021}. The advantage of this model, over other models, was that it was specifically tuned for mathematical language. 

With regards to the input of the model, we followed the preprocessing specified in the previous sections, then we replaced the numerical values that appear in the student response by mask tokens. For example, if we use the response from Llama as an example, the processed response reads as follows:
\begin{quote}
{\em 
One possible way to spend a total of \$\verb_<mask>_ using a combination of chocolates, lollipops, and gum sticks would be to purchase \verb_<mask>_ bags of chocolates ($\$\verb_<mask>_ \times \verb_<mask>_ = \$ \verb_<mask>_$), \verb_<mask>_ bag of lollipops ($\$ \verb_<mask>_$), and \verb_<mask>_ bags of gum sticks ($\$\verb_<mask>_ \times \verb_<mask>_ = \$ \verb_<mask>_$). This combination would total \$\verb_<mask>_.
}
\end{quote}
The masked values are stored in a list for later use. In this particular example, each of the values we want (9,2, and 1) appear twice. In training the models as a token classifier model, the tokens masking the correct values identified hand-scoring team are given target labels of 1 while every other token is given a value of 0. 

One of the pitfalls of this method is that there are occasions where the value coincides with the correct value, but in an incorrect context. For example, if the quantities specified are all the same, this would mean the target label of 1 is applied even in contexts where different values are being specified. The other pitfall is that we often found that the value identified did not actually appear in the student text. In the context of the above example, the student may specify bags of chocolates in two different places and make it clear that the final answer is the addition of the two numbers. There are many different ways in which this may happen. In Appendix \ref{app:full}, we  provide the number of times the value specified by the hand-scoring team did not appear in the student response.

In the validation process, for a particular response, the token classification gives each masked value a probability of being correct. Taking the value associated with the maximal probability defines our choice. The normalization so that the sum of probabilities is 1 does not affect the choice of value. This defines the accuracy for the fine-tuned ELECTRA model and the Math-BERT model.

While we only have two models, it is useful to define the results for an ensemble of a collection of models, $f_1, \ldots, f_m$. For any particular response with values $v_1,\ldots, v_n$, we define $p_{i,j}$ as the probability that model $i$ assigns to value $v_j$ ($i \in \{1,\ldots, m\}$ and $j \in \{1,\ldots, n\}$). If $\alpha_1, \ldots, \alpha_m$ is a collection of weights such that 
\[
0\leq \alpha_k \leq 1 \,\,\text{and}\,\,\sum \alpha_k = 1,
\]
then we can define the ensemble as the linear combination
\[
F(\alpha_1, \ldots, \alpha_m)  = \sum_{i} \alpha_i f_i.
\]
The ensemble probability, $P_j$, assigned to value $v_j$ is given by
\[
P_j = P_j(\alpha_1, \ldots, \alpha_m) = \sum_{i =1}^m \alpha_i p_{i,j}.
\]
The maximum value of $P_j$ defines which value is chosen. 

This means that the accuracy of the ensemble is a function of $\alpha_1,\ldots, \alpha_m$. We can define the ensemble, $\tilde{F}$, as the linear combination that maximizes the accuracy on the development set. Because the accuracy is not a differentiable function, when we defined the ensemble of the ELECTRA and Math-BERT model, we used an adaptation of Powell's method to determine optimal values the $\alpha$ parameters \cite{powell_efficient_1964}. This means that, on the development set, the accuracy of $\tilde{F}$ is greater than, or equal to, the accuracy of each model. This does not mean that the accuracy is necessarily maximized on the test set.

This optimization can be done at a prompt level so that the ensemble is tuned to each prompt and value. By defining an ensemble in this way, this avoids the problem of not having sufficient training data to fine-tune an entire language model. 

\section{Results}\label{sec:results}

As discussed, the results are characterized by four values, $\kappa_0$, $\kappa_1$, $\kappa_v$ and $p$. The final engine is the result of a single classification model, given by ELECTRA, and the individual prompt level models, given by the ensembles of a generic ELECTRA model and a Math-BERT model. The performance statistics are summarized in Table \ref{tab:final_summary}. We have included the results of the Math-BERT model and the ELECTRA model to highlight the benefits of ensembling at a prompt level. A more detailed account of the results at the prompt and value level is left to Appendix \ref{app:full}.

\begin{table*}
\begin{center}
\begin{tabular}{c | c c c  c|  c c c | c c c} \toprule
 & \multicolumn{4}{c}{Test IRR} & \multicolumn{3}{c}{Engine $\kappa$} & \multicolumn{3}{c}{Engine $p$} \\ 
 & $\kappa_0$ & $\kappa_1$ & $\kappa_v$ & $p$ & $\kappa_0$ & $\kappa_1$ & $\kappa_v$ & M-BERT & ELECTRA &  $\tilde{F}$ \\ \midrule
1 & 0.883 & 0.903 & 0.870 & 0.808 &  0.897 & 0.865 & 0.912 & 0.502 & 0.710 & 0.719 \\
2 & 0.909 & 0.864 & 0.920 & 0.940 &  0.897 & 0.865 & 0.912 & 0.890 & 0.897 & 0.911 \\
3 & 0.786 & -  & 0.786  & 0.857 &  0.800 & -       & 0.800 & 0.955 & 0.955 & 0.955 \\
4 & 0.836 & -  & 0.837 & 0.931 &  0.850 & -        & 0.833 & 0.927 & 0.927 & 0.927 \\
5 & 0.708 & 0.888 & 0.721 & 0.918 &  0.743 & 1.000 & 0.748 & 0.964 & 0.971 & 0.971 \\
6 & 0.892 & 0.897 & 0.878 & 0.877 &  0.910 & 0.888 & 0.873 & 0.877 & 0.883 & 0.886 \\
7 & 0.935 & 0.945 & 0.937 & 0.916 &  0.911 & 0.871 & 0.908 & 0.765 & 0.689 & 0.825 \\ \midrule 
Total & 0.850 & 0.899 & 0.850 & 0.891& 0.858 & 0.898 & 0.855 & 0.789 & 0.800 & 0.851 \\ 
\bottomrule
\end{tabular}
\end{center}
\caption{The performance of the engine is presented next to the corresponding statistics that govern the inter-rater reliability statistics. The classification model is evaluated using Cohen's kappa statistics, while the identification model is evaluated by considering the probability of the exact agreement between scores. \label{tab:final_summary}}
\end{table*}

We see that the classification model presented gives an average performance that is on-par with human agreements. This signifies that the pipeline's ability to distinguish when a value is 0, 1, or a separate value specified by the student is at an acceptable level. The second component, where the pipeline discerns where in the text the student has stated the value, is still not at human level performance. It seems like the identification models struggle the most for prompts 1 and 7, where there are a larger number of values to extract. These two prompts also had a large number of values that did not have a large training set for the ensemble. It may also be possible to classify these instances separately.

As an anecdotal test, we subjected the response given by the Llama 2 model to the generic pipeline, where the questions asked were how many bags of each of the items. The classification component on the pipeline indicated that values were specified in each case and the value identification component returned 9 bags of chocolates, 1 bag of lollipops, and 2 bags of gum sticks. Furthermore, when we changed the wording to ``a bag of lollipops", the output was the same, however, the classification component of the pipeline for lollipops returned a value of 1. While this is not indicative of any statistical agreement, it is useful to know that this procedure could be applied in situations beyond the set of problems the generic models were trained on.

\section{Discussion}\label{sec:conclusion}

This paper contains a blueprint for the implicit evaluation of a certain class of short answer items in mathematics. Even though the performance demonstrated in this paper is not at or above human level performance, the use of language models for this particular dataset achieved a level of success that would be acceptable in many circumstances. We believe this method could be adapted to other datasets, and hence, we could adapt this work to provide valuable feedback to teachers and students.

The above is a problem with a wide applicability in a range of quantitative fields. In our case, the application is mathematics in that the aim of this pipeline is to extract key numerical values in text, however, this could also be applied to physics, biology, or chemistry. More generally, beyond educational applications, this pipeline could be useful in data-mining, digital commerce, and other fields where knowledge of numerical values in text is important. 

Many people have considered very large general-purpose language models to as the be-all and end-all of natural language processing. We do not adhere to this belief. It is important to realize that even very small models, using the right architectures and fine-tuned in specific ways to specific tasks, can be useful in ways that large general-purpose models can not at this time. Indeed, there are many cases where smaller models are outperforming large general purpose models. 

\bibliographystyle{plain}
\bibliography{lib}{}

\appendix

\section{Full results}\label{app:full}

In this section, we break down the results for each value and for each prompt. The full list of results for each prompt and each value is provided in Table \ref{tab:full}. 

\begin{table*}[!ht]
    \centering
\begin{small}
\begin{tabular}{l  l r r  | r r r  r}
\toprule
Prompt &  Value & $N$ & $M$ & IRR & M-BERT & ELECTRA & Ensemble \\ \midrule
1& all  & 1144 &  48 & 0.808 & 0.502 & 0.710 & 0.719 \\
& 1      & 35  &   4 & 0.400 & 0.543 & 0.543 & 0.571 \\
& 2      & 28  &   2 & 0.286 & 0.357 & 0.500 & 0.429 \\
& 3      & 27  &   2 & 0.407 & 0.296 & 0.630 & 0.444 \\
& 4      & 78  &   4 & 0.897 & 0.654 & 0.603 & 0.590 \\
& 5      & 98  &  11 & 0.786 & 0.245 & 0.571 & 0.571 \\
& 6      & 95  &  14 & 0.821 & 0.421 & 0.716 & 0.716 \\
& 7      & 253 &   4 & 0.866 & 0.787 & 0.719 & 0.787 \\
& 8      & 252 &   2 & 0.861 & 0.381 & 0.706 & 0.706 \\
& 9      & 278 &   5 & 0.827 & 0.457 & 0.831 & 0.831 \\ \midrule
2 & all & 619  &   14 & 0.940 & 0.890 & 0.897 & 0.911 \\
 &   1  & 99   &   8 & 0.970 & 0.848 & 0.818 & 0.848 \\
 &   2  & 37   &   1 & 1.000 & 0.919 & 0.865 & 0.919 \\
 &   3  & 483  &   5 & 0.930 & 0.896 & 0.915 & 0.923 \\ \midrule
3 & all & 265  &   3 & 0.857 & 0.955 & 0.955 & 0.955 \\ \midrule
4 & all & 438  &  10 & 0.932 & 0.927 & 0.927 & 0.927 \\ \midrule
5 & all & 476  &  10 & 0.918 & 0.964 & 0.971 & 0.971 \\ \midrule
6 & all & 1591 &  95 & 0.877 & 0.877 & 0.883 & 0.886 \\ 
    & 1 & 136  &   6 & 0.882 & 0.890 & 0.904 & 0.904 \\
    & 2 & 82   &  13 & 0.939 & 0.756 & 0.793 & 0.793 \\
    & 3 & 137  &  14 & 0.883 & 0.818 & 0.803 & 0.803 \\
    & 4 & 114  &  19 & 0.807 & 0.746 & 0.746 & 0.746 \\
    & 5 & 197  &  15 & 0.883 & 0.848 & 0.858 & 0.858 \\
    & 6 & 131  &  10 & 0.908 & 0.870 & 0.893 & 0.893 \\
    & 7 & 69   &   9 & 0.826 & 0.783 & 0.812 & 0.812 \\
    & 8 & 725  &   9 & 0.877 & 0.939 & 0.938 & 0.943 \\ \midrule
7 & all & 2412 &  13 & 0.917 & 0.765 & 0.689 & 0.825 \\
    & 1 & 379  &   0 & 0.921 & 0.879 & 0.462 & 0.897 \\
    & 2 & 36   &   2 & 0.917 & 0.583 & 0.556 & 0.556 \\
    & 3 & 161  &   1 & 0.907 & 0.919 & 0.876 & 0.907 \\
    & 4 & 144  &   0 & 0.875 & 0.771 & 0.722 & 0.778 \\
    & 5 & 32   &   1 & 0.812 & 0.312 & 0.781 & 0.781 \\
    & 6 & 362  &   0 & 0.928 & 0.843 & 0.873 & 0.903 \\
    & 7 & 43   &   2 & 0.860 & 0.791 & 0.953 & 0.953 \\
    & 8 & 371  &   4 & 0.927 & 0.844 & 0.946 & 0.960 \\
    & 9 & 227  &   2 & 0.925 & 0.758 & 0.793 & 0.824 \\
    & 10 & 215 &   0 & 0.921 & 0.749 & 0.172 & 0.749 \\
    & 11 & 215 &   0 & 0.921 & 0.195 & 0.260 & 0.265 \\
    & 12 & 227 &   1 & 0.916 & 0.855 & 0.956 & 0.960 \\
\bottomrule
\end{tabular}
\end{small}
    \caption{The prompt number and value number appear in the first two columns, the values in the column labeled N are the total number of values not equal to 0 or 1 and the column labeled M contains the number of elements missing after the text was preprocessed. The remaining columns contain the agreements between the raters, in the case of IRR, and the agreement rate between the final value and the engines for the other columns.}
    \label{tab:full}
\end{table*}

Table \ref{tab:full} gives us some very useful insights. One of the insights we have was that there was possibly a need for more preprocessing to minimize the number of missing values. For example, while the second value of prompt 6 had an agreement of only 0.746 with the final score, given 13 out of 82 values did not appear, then the maximal accuracy that was possible from this pipeline was only 0.841. 

There were also instances where the pipeline categorically failed, such as value 11 of prompt 7. In the same way, there were also instances, like prompt 1, values 1 to 3, where there clearly wasn't very high agreement between human raters. There were also many occasions where the agreement statistics for the models exceeded the agreements between raters. Notably with the one value prompts (3,4, and 5),  and some cases where there were a large number of training examples. 

Clearly, the ensemble improved the overall performance by a significant margin. There were a total of 5 cases out of 35 in which the ensemble performed worse on the test set than one of the models. In these cases, the ensemble, $\tilde{F}$, still performed better on the development set, but not the test set. 

Overall, the agreement between the humans was above the agreement of the ensemble by about 5\%. Even so, this discrepancy is usually considered acceptable for production purposes. We believe ensmbling more models could improve the performance, however, to simplify this exposition, we have limited our ensemble to just two models. 
\end{document}